\documentclass[letterpaper, 10 pt, conference]{ieeeconf}  
\usepackage{array,booktabs}
\usepackage{mathtools, nccmath}

\DeclareMathOperator{\diag}{diag}
\usepackage{dsfont}

\IEEEoverridecommandlockouts

\usepackage{graphicx}
\usepackage{tikz}
\usetikzlibrary{shapes.geometric}
\usepackage{pgfplots}
\usetikzlibrary{patterns}
\usepackage{textcomp}
\usepackage{siunitx}
\usepgfplotslibrary{groupplots,dateplot}
\usetikzlibrary{patterns,shapes.arrows}
\pgfplotsset{compat=newest}
\usepackage{adjustbox}
\usepackage{multirow}
\usepackage{subcaption}
\usepackage[hidelinks]{hyperref}

\usepackage{pifont}
\definecolor{ForestGreen}{RGB}{34,139,34}
\definecolor{GoldenRod}{HTML}{FFDF42}
\definecolor{DeepBlue}{HTML}{2D2F92}
\newcommand{\Vmark}{\textcolor{black}{\ding{51}}}%

\newcommand{\DDtwo}{$\mathcal{D}_{\textbf{\texttt{DD2}}}$}
\newcommand{\DG}{$\mathcal{D}_{\textbf{\texttt{G}}}$}

\newcommand{\Dreal}{$\mathcal{D}_{\textbf{\texttt{real}}}$}

\usepackage{amsmath} 
\usepackage{amssymb}  
\usepackage{verbatim}
\usepackage{rotating}

\title{\LARGE \bf R2SNet: Scalable Domain Adaptation for Object Detection in Cloud--Based Robotic Ecosystems via Proposal Refinement}

\author{Michele Antonazzi, Matteo Luperto, N. Alberto Borghese, Nicola Basilico
\thanks{All authors are with the Department of Computer Science, University of Milan, Milano, Italy {\tt\small~name.surname@unimi.it}}%
}

\begin{document}

\maketitle
\thispagestyle{empty}
\pagestyle{empty}

\begin{abstract}
We introduce a novel approach for scalable domain adaptation in cloud robotics scenarios where robots rely on third--party AI inference services powered by large pre--trained deep neural networks. Our method is based on a downstream proposal--refinement stage running locally on the robots, exploiting a new lightweight DNN architecture, R2SNet. This architecture aims to mitigate performance degradation from domain shifts by adapting the object detection process to the target environment, focusing on relabeling, rescoring, and suppression of bounding--box proposals. Our method allows for local execution on robots, addressing the scalability challenges of domain adaptation without incurring significant computational costs. Real--world results on mobile service robots performing door detection show the effectiveness of the proposed method in achieving scalable domain adaptation. 
\end{abstract}

\section{Introduction}\label{sec:intro}

Robot--assisted services are today present in a wide range of real--world applications, including healthcare, logistics, domestic assistance, and agriculture~\cite{alatise2020review}. While becoming more and more ubiquitous, autonomous mobile robots are facing a growing need to tackle increasingly complex perception and decision--making tasks for which the recent wave of AI and deep learning offers solutions of unprecedented potential, often available as very large Deep Neural Networks (DNNs) pre--trained on public or third--party datasets.

The computational capabilities that such a need brings are at odds with the typical profiles of mobile robots: not only are they devices with limited resources, but they need to be. Keeping affordable hardware costs and preserving energy consumption at operational time are mandatory requisites in many real--world scenarios. This is the reason why offloading the computationally demanding inference with DNNs is an emerging trend in the field, for which third--party AI services deployed in the cloud are a convenient solution. Such services have great capabilities, but, as many robotic practitioners are well aware of, also have access constraints and performance barriers. Constraints typically entail that they can only be accessed with queries. Among performance limiting factors, domain shifts are perhaps the most relevant to the field--AI paradigm that robots embody: the data distribution encountered in their target environments can significantly diverge from the distribution on which the cloud--based DNN has been trained. This discrepancy can inevitably result in substantial performance degradation.

\begin{figure}[t]
    \centering
    \includegraphics[width=\linewidth]{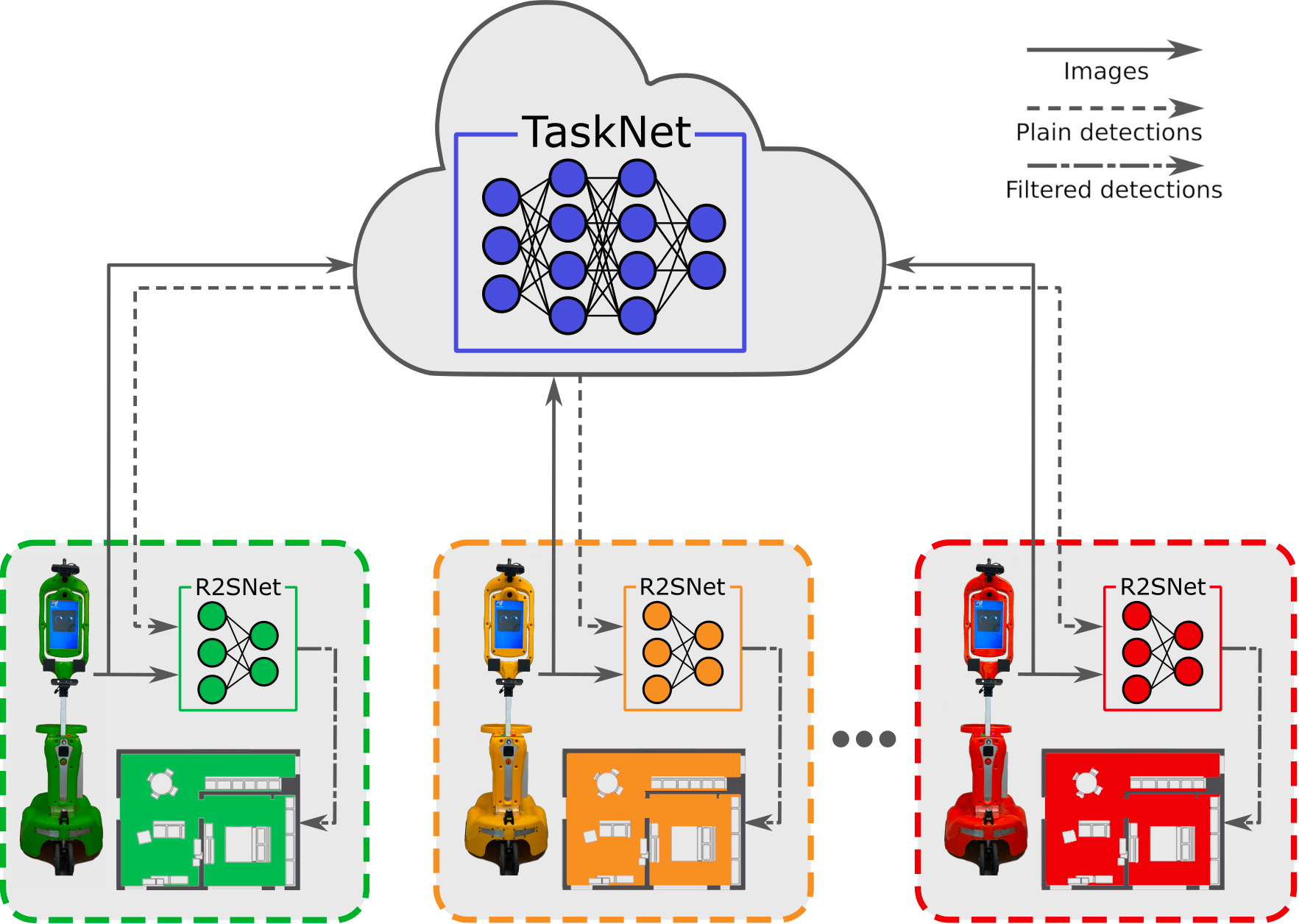}
    \caption{A general overview of the cloud--based scenario we consider.}
    \label{fig:ecosystem}
\end{figure}

Consider these challenges in the scope of a robotic ecosystem where multiple independent units are deployed across different environments and rely on a cloud--based DNN inference service. Assume that the robots' working environments are initially unknown and the number of robots in the system is expected to increase by deploying new units in novel environments. Standard \emph{domain adaptation} techniques~\cite{oza2023unsupervised, lee2022surgical}, where fine--tuning and feature alignment are exploited to train models that can withstand the shift to a target domain, face scalability issues in such a scenario. The fields of Cloud Robotics~\cite{hu2012cloud}, and more recently Fog Robotics~\cite{tanwani2019fog}, come in handy by studying inference--serving solutions that distribute, in an adaptive way, the computational and storage loads across robots and cloud services. 
However, the application of domain adaptation techniques over these architectures is subject to scalability issues. First, it requires full access to the cloud DNN; this is not always feasible if the cloud DNN is provided by an external vendor. Then, performing domain adaptation on the full DNN model would pose significant costs; each time a new robot is deployed, a new DNN is to be trained, deployed, and maintained, as it cannot be shared by multiple robots after performing domain adaptation.

In this work, we focus on scalable domain adaptation, a problem at the intersection of cloud robotics and deep learning that, despite being relevant to many real--world settings, is still largely underexplored. We focus on the task of object detection in the general scenario represented by Fig.~\ref{fig:ecosystem}: a set of robots need to carry out such a task from RGB images acquired in their respective environments. To such end, they rely on a general--purpose pre--trained DNN, called here TaskNet, which is provided as a third--party cloud service, making it accessible exclusively through queries.

The core contribution of our approach is to perform domain adaptation as an efficient downstream proposal--refinement stage, running locally on the robots. As we shall detail in Section~\ref{sec:proposals_refinement}, this strategy is inspired by the observation that state--of--the--art object detectors typically work by generating dense sets (up to thousands) of bounding--box proposals which then undergo heuristic post--processing via confidence thresholding and non--maximum suppression~\cite{hosang-learning-nms}. Our findings indicate that a substantial portion of the performance degradation due to domain shifts can be mitigated by introducing before such post--processing heuristics a proposal--refinement step adapted to the target environment. To such end, we introduce R2SNet, a novel lightweight DNN architecture for proposal refinement that focuses on three different types of corrective actions: relabeling, rescoring, and suppression of bounding boxes. To carry out such a task, R2SNet leverages the acquired images and the geometrical features of the corresponding bounding--box proposals, and it can be run downstream and locally on the robot.

We evaluate this method in a real--world testbed where mobile service robots must perform real--time \emph{door detection}, that is identifying the location and status (open/closed) of doors/passages through visual recognition~\cite{antonazzi2024development}. For service robots, this object detection task is key for navigation, but also one recognized as very much affected by domain shifts~\cite{antonazzi2023enhancing}. The obtained results show how our method enables scalable adaptation, effectively mitigating the performance losses due to domain shifts encountered with the general pre--trained model, all while avoiding the need for substantial computational costs in training and inference.
\section{Related Works}\label{sec:realated_works}

Cloud Robotics~\cite{hu2012cloud} is an active area of research focusing on engineering the distribution of storage and computational tasks away from robotic platforms to web--enabled architectures~\cite{afrin2021resource}. In the last years, this area faced the wave of cyber--physical systems' increasing reliance on large, pre--trained DNNs. Such models pose substantial computational demands that fostered the development of strategies for distributing their workload to the edge, towards ecosystems where edge computing and deep learning become interlinked~\cite{liu2023roboec2}. Object detection represents one of the most significant testbeds against this background~\cite{wang2020convergence},~\cite{guo2019distributed}.

One of the mainstream approaches for cloud--based DNN workload distribution is Model Splitting~\cite{abuadbba2020can} where, essentially, the model is divided into two or more portions that are run collaboratively across the network. Examples of this method for object detection are~\cite{chakroun2021distributing} where YOLOv3 undergoes a process of cloud--edge distribution and~\cite{teerapittayanon2017distributed} where inference follows a hierarchical structure from the cloud to the end device. 

A series of works, falling under the umbrella of ``Fog Robotics'', investigated solutions based on a continuum of computing resources from robots to cloud data centers. This paradigm is becoming increasingly widespread~\cite{FogROS2}, with distributed object detection (typically in synergy with grasping) being among its real--world challenges. Examples of works on this line include~\cite{tanwani2019fog}, where models are initially trained in the cloud and subsequently adapted at the edge,~\cite{tanwani2020rilaas} where authors focus on reducing latency by means of a Q--learning--based policy for load balancing, and~\cite{vinod2022development} where object detection based on SSD~\cite{liu2016ssd} is served to the robot from a fog node cluster whose resources can be adapted to guarantee service quality. Other examples of similar offloading strategies for object detection in mobile robots have been proposed in~\cite{beksi2015core} and~\cite{penmetcha2020smart}. 

Most of these works primarily focus on enhancing service quality but overlook the challenge of scalable domain adaptation in the constrained cloud setting we adopt. In this paper, we directly address this issue with an architecture related to the one proposed in~\cite{chinchali2021network} where cooperation between a large cloud--based model and a smaller one operating locally on the robot is exploited. Our method differentiates in the role and design of the smaller model. In such work, the smaller model is essentially a scaled--down, less accurate variant of the larger one, aimed at reducing costs. 
In contrast, our approach enhances the smaller model's role to not just serve as a cost--effective alternative but to specifically refine and adapt the cloud model's predictions for the robot's unique operational environment in a scalable way.

\section{Method}


\subsection{Proposals Filtering in Object Detection}\label{sec:proposals_refinement}

Object Detection (OD) amounts to identifying the location and dimension of objects in an image. 
Deep learning is today the leading approach to building detectors, which are typically based on architectures that analyze the input image through different stages to ensure its comprehensive coverage~\cite{objectdetectionsurvey}. Two--stage detectors (such as Faster R--CNN~\cite{fasterrcnn}) use a Region Proposal Network (RPN) to predict, in a first stage, proposals of bounding boxes from multi--scale image embeddings. In the second stage, such proposals are classified into object categories. Differently, one--stage models (such as YOLO~\cite{yolov3}) directly predict object classes for a set of predefined bounding boxes called \emph{anchors}, which uniformly cover the image with multiple scales and sizes.

Both architectures share a characteristic: they produce many overlapping bounding--box proposals, typically numbering in the thousands, which are independently scored using the image's features. To distill meaningful detections from this dense set, a heuristic two--step post--processing is commonly executed. The first step, called Non--Maximum Suppression (NMS)~\cite{hosang-learning-nms}, iteratively selects pairs of proposals whose Intersection over Union area (IoU) exceeds a threshold $\rho_{IoU}$ and suppresses the one with the lowest confidence. 
In the second step, any proposal with a confidence lower than a threshold $\rho_c$ is also discarded.



For achieving scalable adaptation, we suggest relocating the post--processing step to operate locally on each robot. This entails integrating it as a downstream module of a global cloud--based object detector we call TaskNet (see Fig.~\ref{fig:ecosystem}) which has been configured to return raw proposals by modifying hyperparameters. Additionally, we augment this post--processing by incorporating R2SNet, a lightweight deep architecture tailored to the robot's target environment.


From an image $x$, we obtain from the TaskNet a set of raw proposal \mbox{$\hat{Y} = \{\hat{y}\}$}, with
\begin{equation}
    \hat{y} = \big[\hat{c}_x, \hat{c}_y, \hat{w}, \hat{h}, \hat{c}, \text{hot}(\hat{o})\big]_{1 \times \hspace{-0.04em}f}\label{eq:bbox_predicted}
\end{equation}
where $\hat{c}_x, \hat{c}_y \in [0, 1]$ are the center coordinates,  $\hat{w}, \hat{h} \in [0, 1]$ represent width and height, $\hat{c}$ is the confidence, $\hat{o} \in \mathcal{O}$ is an integer indicating the object category, and $\text{hot}(\cdot)$ is its one--hot encoding (so $f = 5 + |\mathcal{O}|$). Once received by the robot, the $k$ most confident proposals, where $k \gg O$, with $O$ indicating the maximum number of identifiable objects in an image, are given as input to R2SNet. Before presenting its architecture, we examine the three primary types of interventions along which the network is trained and used: Relabeling, Rescoring, and Suppression (hence the acronym R2SNet).
In the remainder of the paper, we focus on a specific object detection task, \emph{door detection}, as it is particularly significant for this task, as discussed in the examples below.
However, our considerations are general to other detection tasks for autonomous robots.

\begin{figure}[t]
	\centering
    \begin{tabular}{@{}c@{ }c@{ }c@{ }c@{ }}
    \rotatebox[origin=c]{90}{\scriptsize a -- Relabeling}&
    \raisebox{-0.45\height}{\includegraphics[width=0.31\linewidth]{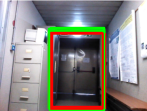}}&
    \raisebox{-0.45\height}{\includegraphics[width=0.31\linewidth]{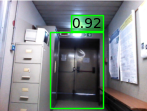}}&
    \raisebox{-0.45\height}{\includegraphics[width=0.31\linewidth]{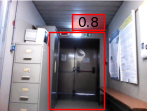}}\\\addlinespace[0.1cm]
    \rotatebox[origin=c]{90}{\scriptsize b -- Rescoring}&
    \raisebox{-0.45\height}{\includegraphics[width=0.31\linewidth]{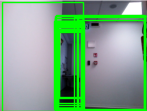}}&
    \raisebox{-0.45\height}{\includegraphics[width=0.31\linewidth]{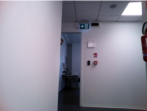}}&
    \raisebox{-0.45\height}{\includegraphics[width=0.31\linewidth]{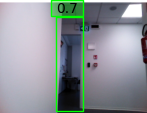}}\\\addlinespace[0.1cm]
    \rotatebox[origin=c]{90}{\scriptsize c -- Suppression}&
    \raisebox{-0.45\height}{\includegraphics[width=0.31\linewidth]{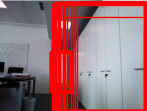}}&
    \raisebox{-0.45\height}{\includegraphics[width=0.31\linewidth]{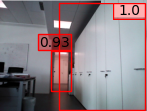}}&
    \raisebox{-0.45\height}{\includegraphics[width=0.31\linewidth]{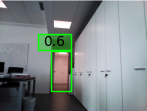}}\\\addlinespace[0.1cm]
    &{\scriptsize TaskNet} & {\scriptsize TaskNet + NMS} & {\scriptsize TaskNet + R2SNet + NMS} 
\end{tabular}
	\caption{R2SNet refinements in filtering dense proposals, compared to standard post--processing. Green/red bounding boxes are open/closed doors.}
	\label{fig:net_goals}
     \vspace{-0.5cm}
\end{figure}

\subsubsection*{Relabeling} 
Frequently, a TaskNet generates several overlapping proposals over the same target object; some of these proposals often have contrasting labels. Fig. 2a shows an example of different overlapping proposals that label the same door both as closed and open.
These errors are frequent when involve objects that might resemble each other (e.g., open and closed doors or chairs and armchairs). The standard post--processing based on NMS would select the proposal with the highest confidence disregarding the correctness of its object category.
In our method, we improve on this by relabeling all overlapping proposals to a single category, forcing a consensus. Also, we identify isolated proposals not overlapping with any others as spurious. Based on our empirical observations, these isolated proposals often correspond to errors in object localization. Consequently, we relabel them as \texttt{background}, an additional category introduced in this stage.

\subsubsection*{Rescoring} 

It is well--known that confidence scores may not consistently reflect the actual uncertainty, and thus the likelihood of correctness, of the proposals computed with TaskNet~\cite{popordanoska_calibration_od}. When a poorly localized proposal receives high confidence, NMS might erroneously reject nearby proposals that better match with the object. Conversely, if a properly localized proposal receives low confidence, the thresholding step could erroneously discard it. We observed this phenomenon in challenging instances, such as the one depicted in Fig.~\ref{fig:net_goals}b, where an open door is partially hidden behind a corner, often leading to errors.
To address this, we correlate the confidence of each proposal to the IoU area they have with the best overlapping ground truth box. In this way, the IoU threshold $\rho_{IoU}$ becomes the only hyperparameter for the post--processing techniques.

\subsubsection*{Suppression}

Other frequent errors occur when dense sets of proposals are situated in parts of the image where no objects are present. This might happen because the features in these regions mimic an object category that the detector is trained to identify. For instance, Fig.~\ref{fig:net_goals}c highlights instances where cabinets (or windows) are misclassified as doors. While relabeling partially mitigates this issue, we introduce a suppression phase to directly address it by learning a feature embedding from the image portion of each proposal to differentiate between background areas and those containing an object.




\subsection{R2SNet}

\begin{figure*}[t]
    \centering
    \includegraphics[width=\linewidth]{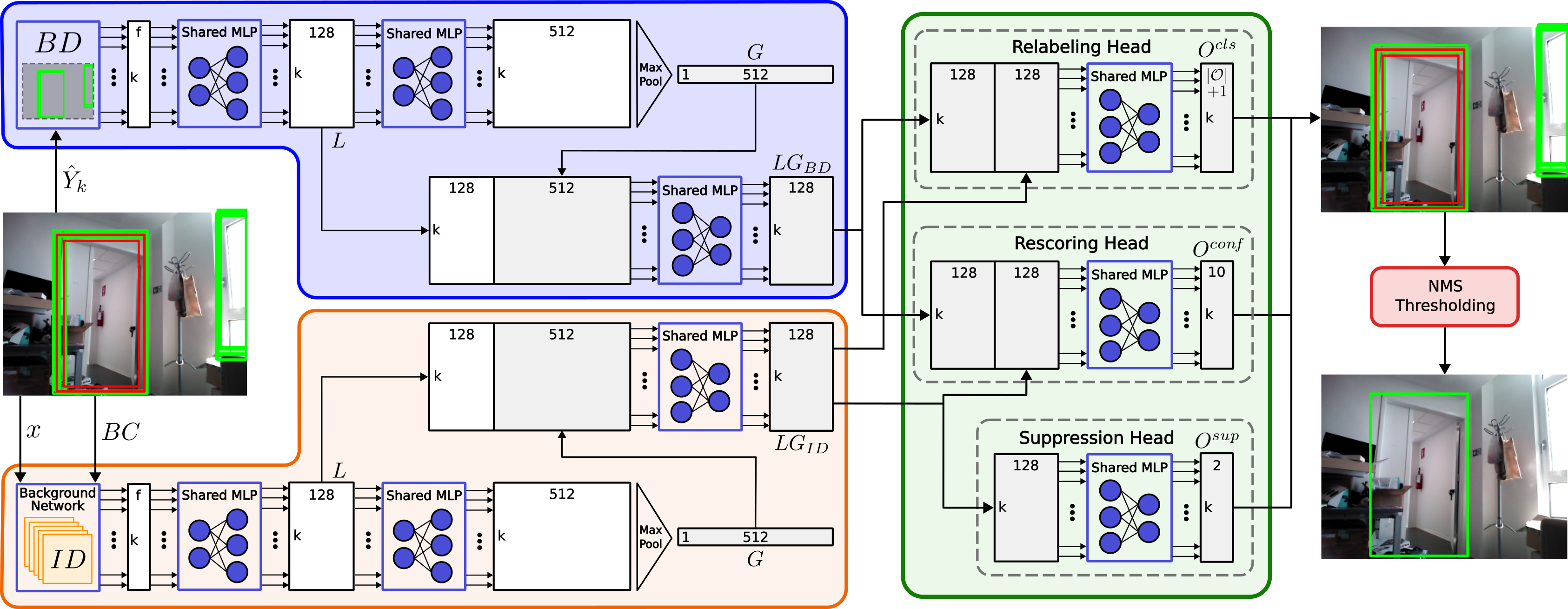}
    \caption{The R2SNet architecture. Batch normalization and ReLU activation functions are applied to all layers of the shared MLPs.}
    \label{fig:R2SNet-general}
        \vspace{-0.4cm}
\end{figure*}





First, given the $k$ most confident proposals computed by the TaskNet, R2SNet extracts a matrix of Bounding--box Descriptors \mbox{$BD = \big[\hat{y}_1, \hat{y}_2, \ldots, \hat{y}_k  \big]_{k \times \hspace{-0.04em}f}$} by stacking the vectors defined in Eq.~\ref{eq:bbox_predicted}. Additionally, it extracts a feature vector from the portion of the image $x$ corresponding to each proposal, using a convolutional architecture we call BFNet (Bounding--box Feature Network, detailed in Sec.~\ref{sec:bfnet}), to compute a matrix of image descriptors \mbox{$ID_{[k \times 8]}$}. \mbox{$BD$} and \mbox{$ID$} can be seen as projections of the $k$ most confident bounding boxes in two distinct spaces $\mathbb{R}^f$ and $\mathbb{R}^8$, which are meant to capture their geometrical and visual features.

Given these preliminaries, R2SNet (depicted in Fig.~\ref{fig:R2SNet-general}) is inspired by PointNet~\cite{pointnet}, which is designed to classify and segment dense point clouds and to be invariant to input permutation (proposals, in our setting). 
R2SNet processes $BD$ and $ID$ with two symmetric sub--networks. At first, each of them maps the input to a high--dimensional space using a Multi--Layer Perceptron (MLP) shared across the $k$ proposal descriptors, to obtain local features $L_{[k \times l]}$. In the MLPs, the same weights are applied to each descriptor, making the size of the network fixed regardless of the number $k$ of proposals. After this step, the local features are expanded again with another MLP and then aggregated using $\max$, to obtain a global feature vector $G_{[1 \times g]}$. This last one is concatenated with each row of $L_{[k \times l]}$ and then mixed with a shared MLP, obtaining an embedding $LG_{[k \times 128]}$ that represents both local and global features of the $k$ proposals.
The outputs $LG_{BD}$ and $LG_{ID}$ of the two sub--networks are fed into three heads to handle the relabeling, rescoring, and suppression of the proposals. 

\begin{figure*}[t]
    \centering
    \includegraphics[width=\linewidth]{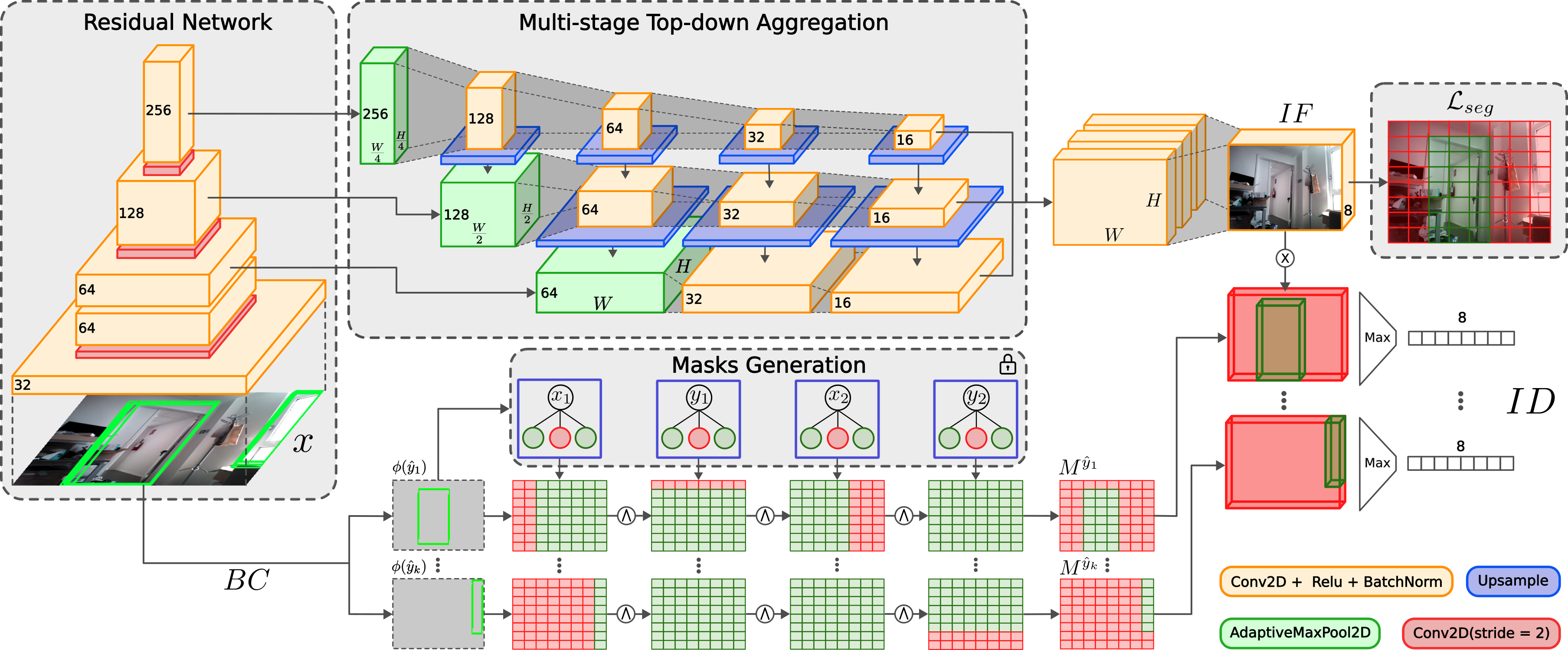}
    \caption{The BFNet architecture.
    }
    \label{fig:R2SNet-background}
    \vspace{-0.4cm}
   \end{figure*}

We denote as $Y$ the set of ground truth bounding boxes for image $x$ where each $y \in Y$ is encoded as per Eq.~\ref{eq:bbox_predicted} by setting $c=1$. We define a matching rule to assign a proposal $\hat{y}$ to a ground--truth bounding box $\hat{y}^{GT} = \arg\max_{y \in Y} a_{IoU}(\hat{y}, y)$, where $a_{IoU} (\hat{y}, y)$ is the IoU area between $\hat{y}$ and  $y$. 

The relabeling head, starting from the concatenation of $LG_{BD}$ and $LG_{ID}$, assigns to each proposal $\hat{y}$ the probabilities for each object class in the set $\mathcal{O} \cup \{\texttt{background}\}$, producing an output  $O^{cls}_{[k \times |\mathcal{O}| + 1]}$. This head is trained with the following log--loss:
\begin{equation}
   \mathcal{L}_{cls}(O^{cls}) = -\frac{1}{k} \sum_{p=1}^{k}\log\big(O^{cls}_p\big) \cdot \text{hot}\big(\hat{o}_p\big),
   \label{eq:loss_relabeling}
\end{equation}
where $(\cdot)$ is the dot product and $\hat{o}_{p}$ is the true class for the $p$-th proposal determined by our matching rule: 

\begin{equation*}
\hat{o}_{p} = \begin{cases}
    Class(\hat{y}^{GT}_p) & \text{if } a_{IoU}(\hat{y}^{GT}_p, \hat{y}_p) \geq \rho_{IoU}\\
    \texttt{background} & \text{otherwise.}
\end{cases}
\end{equation*}


The rescoring head aligns the confidence of a proposal $\hat{y}$ to its IoU area with its associated ground truth $\hat{y}^{GT}$. To achieve this, the confidence score $c\in[0, 1]$ is discretized into $10$ intervals. The rescoring head is then tasked with predicting the likelihood that the confidence score falls within each of these intervals, yielding an output matrix $O^{con\hspace{-0.1em}f}_{[k \times 10]}$. For training, we construct a target vector $v(\hat{y})$ for a proposal $\hat{y}$, whose values peak at the interval corresponding to the IoU score between $\hat{y}$ and its corresponding ground truth $\hat{y}^{GT}$, and decrease in a Gaussian--like manner on either side of the peak. In such a way, we obtain a measure of the error that increases with the distance between the predicted and true peaks. This error is adopted for the rescoring loss:
\begin{equation}
    \mathcal{L}_{res}(O^{con\hspace{-0.1em}f}) = \frac{1}{k}\sum_{p=1}^{k} \|   O^{con\hspace{-0.1em}f}_{p} - v(\hat{y}_p) \|_1.
\end{equation}
The confidence assigned to each $\hat{y}_p$ is $\arg\max_j O^{con\hspace{-0.1em}f}_{p,j}$.

Finally, the suppression head is trained with a loss obtained by adapting Eq~\ref{eq:loss_relabeling} for binary classification between proposals that correspond to an object (those for which $\hat{o}_p \neq \texttt{background}$) and those falling on the background.





\subsection{BFNet}\label{sec:bfnet}

BFNet, shown in Fig.~\ref{fig:R2SNet-background}, guides R2SNet to identify those proposals that are wrongly placed on the background and that can be suppressed. This task is challenging when only the descriptors $BD$ (location, confidence, and class) are used, thus image descriptors $ID$ are needed. BFNet partitions the input image with a low--resolution grid mask $M_{[W \times H]}$. Then it extracts a feature encoding $IF_{[8 \times W \times H]}$ ($8$ channels for each cell), which is mapped to each proposal's region of interest.

More precisely, BFNet extracts a multi--scale feature hierarchy of the input image using a CNN--based backbone with residual connections~\cite{resnet}. The last three embeddings are re--scaled with dimensions $[W \times H]$, $\big[\frac{W}{2} \times \frac{H}{2}\big]$, and $\big[\frac{W}{4} \times \frac{H}{4}\big]$ using adaptive average pooling layers. To aggregate features at different scales, Feature Pyramid Networks~\cite{fpn} (FPN) are commonly used. Differently from what is done in FPNs, to have more descriptive features,  each embedding is processed by three parallel convolutional backbones and step--by--step top--down aggregated through upsampling and summation. The resulting embeddings are concatenated and mixed through convolution to generate a feature map $IF$.

Then, we need to obtain the portion of $IF$ covered by each proposal. Rather than iteratively slicing $IF$ according to each bounding box coordinates, we perform a faster parallel end--to--end mask generation process. More precisely, we use a series of MLPs that produces a binary mask $M_{[W \times H]}^{\hat{y}}$ for each proposal $\hat{y}$ where an element is set to 1 (0) if inside (outside) the area of $\hat{y}$. This mask is used to suppress the features of $IF$ exceeding the bounding box's boundaries. First, BFNet receives in input a matrix \mbox{$BC = [\phi(\hat{y}_1) \ldots \phi(\hat{y}_k)]_{k \times 4}$}, where $\phi : \mathbb{R}^f \rightarrow \mathbb{N}^4$ encodes an input proposal $\hat{y}$ to a vector $[x_0, y_0, x_1, y_1]$ containing the coordinates of the bottom--left and top--right corners in the grid mask $M_{[W \times H]}$.  It then computes, for each proposal $\hat{y}$, four binary grids defined as
\begin{align}
\begin{split}
    &M^{x_j} =\scalebox{1.2}{$\mathds{1}$}_{\scalebox{0.7}{$\leq$}}\Big((-1)^j(x_j - A)\Big)\\
    &M^{y_j} = \scalebox{1.2}{$\mathds{1}$}_{\scalebox{0.7}{$\leq$}}\Big((-1)^j(y_j - B^T)\Big)
\end{split}\label{eq:masks}
\end{align}
for $j\in\{0,1\}$, where $A = \diag(I_H) \times [0 \ldots W-1]$ and $B = \diag(I_W) \times [H-1 \ldots 0]$. 
We obtain the matrices using four MLPs, each comprising one input and $W \times H$ output neurons. The weights are initialized according to Eq.~\ref{eq:masks} and remain fixed during the training process. The mask of each proposal $\hat{y}$, obtained as 
\begin{equation*}
    M^{\hat{y}}_{[W \times H]} = \bigwedge\limits_{j= 0}^{1}M^{x_j} \wedge M^{y_j},
\end{equation*}
is combined with the embedding $IF$ to suppress the features outside the bounding box boundaries. The results are then compressed along the last two dimensions with a $\max$ operation, obtaining $ID_{[k \times 8]}$ that encodes the image descriptors of each proposal for R2SNet.

Before training the whole R2SNet, BFNet is pre--trained for addressing a low--resolution binary segmentation task. The image features $IF_{[8 \times W \times H]}$ are convoluted into a binary grid mask $M^{seg}_{[2 \times W \times H]}$ obtained by training on this loss:

\begin{equation}
    \mathcal{L}_{seg}(M^{seg})  = -\frac{1}{WH} \sum_{w=1}^W \sum_{h=1}^H \log\big(M^{seg}_{w, h}\big) \cdot \text{hot}\big(l_{w, h}\big),
\end{equation}

where the ground truth label for each cell $l_{w, h}$ is 

\begin{equation*}
    l_{w, h} = 
    \begin{cases}
        1 & \text{if } \exists~y \in Y \mid \phi(y) \text{ contains the cell } w, h  
        \\
         0  & \text{otherwise. } 
    \end{cases}
\end{equation*}

\begin{table*}[t]

\setlength\tabcolsep{4.5pt}
\setlength\extrarowheight{2pt}
\centering
\begin{adjustbox}{width=2.05\columnwidth,center}
\begin{tabular}{c|c@{\hskip 0.1cm}c@{\hskip 0.2cm}c@{\hskip 0.1cm}c@{\hskip 0.1cm}|c@{\hskip 0.1cm}c@{\hskip 0.1cm}c@{\hskip 0.1cm}c@{\hskip 0.1cm}|c@{\hskip 0.1cm}c@{\hskip 0.1cm}c@{\hskip 0.1cm}c@{\hskip 0.1cm}|c@{\hskip 0.1cm}c@{\hskip 0.1cm}c@{\hskip 0.1cm}c@{\hskip 0.1cm}|c@{\hskip 0.1cm}c@{\hskip 0.1cm}c@{\hskip 0.1cm}c@{\hskip 0.1cm}}
\toprule
         &  \multicolumn{4}{c|}{$e_1$} & \multicolumn{4}{c|}{$e_2$} &\multicolumn{4}{c|}{$e_3$}&\multicolumn{4}{c|}{$e_4$}&\multicolumn{4}{c}{$\overline{e}$} \\
         \textbf{Exp.} & \textbf{mAP}$\uparrow$& $\mathbf{TP}$$\uparrow$ &  $\mathbf{FP}$$\downarrow$ & $\mathbf{BFD}$$\downarrow$ & \textbf{mAP}$\uparrow$& $\mathbf{TP}$$\uparrow$ &  $\mathbf{FP}$$\downarrow$ & $\mathbf{BFD}$$\downarrow$ & \textbf{mAP}$\uparrow$& $\mathbf{TP}$$\uparrow$ &  $\mathbf{FP}$$\downarrow$ & $\mathbf{BFD}$$\downarrow$ & \textbf{mAP}$\uparrow$& $\mathbf{TP}$$\uparrow$ &  $\mathbf{FP}$$\downarrow$ & $\mathbf{BFD}$$\downarrow$ & \textbf{mAP}$\uparrow$& $\mathbf{TP}$$\uparrow$ &  $\mathbf{FP}$$\downarrow$ & $\mathbf{BFD}$$\downarrow$ \\
\midrule
TaskNet&33 &41\% & 10\%  & 13\%&27 &33\% & 5\%  & 18\%&13 &24\% & 7\%  & 34\%&47 &47\% & 6\%  & 14\%&30 &36\% & 7\%  & 20\%\\
$\text{R2S}^{30}_{25}$&39 &50\% & 7\%  & 11\%&30 &36\% & 4\%  & 10\%&20 &26\% & 7\%  & 12\%&58 &64\% & 4\%  & 11\%&37 &44\% & 6\%  & 11\%\\
$\text{R2S}^{30}_{50}$&40 &49\% & 8\%  & 9\%&35 &40\% & 6\%  & 6\%&22 &29\% & 7\%  & 10\%&59 &63\% & 5\%  & 9\%&39 &45\% & 6\%  & 9\%\\
$\text{R2S}^{30}_{75}$&49 &54\% & 5\%  & 7\%&35 &39\% & 6\%  & 6\%&26 &31\% & 8\%  & 10\%&62 &62\% & 1\%  & 5\%&43 &46\% & 5\%  & 7\%\\
\bottomrule
\end{tabular}
\end{adjustbox}
\caption{R2SNet performance evaluation when trained with an increasing amount of data. $\overline{e}$ is the average.}\label{tab:data_quantity}
\vspace{-0.5cm}

\end{table*}
\section{Experimental Evaluation}

\subsection{Experimental Setting}

Training is performed using 2 publicly available datasets (see~\cite{antonazzi2024development, antonazzi2023enhancing} for more details) for door detection in RGB images. The first one, \emph{DeepDoors2} (\DDtwo) \cite{deepdoors2}, has 3K real--world images containing doors, taken from a human perspective. The second one, \DG{}, is taken in 10 environments of Gibson~\cite{gibson} and contains around 5K photorealistic images acquired from the viewpoint of a mobile robot. The experimental evaluation is performed on a third, real--world, dataset, \Dreal, collected with our Giraff--X robot (Fig.~\ref{fig:ecosystem})~\cite{giraff}; it contains four runs collected when different indoor environments $e_*$ (three university facilities and an apartment, see Fig.~\ref{fig:net_goals}) have been fully mapped. All datasets have \mbox{$\mathcal{O} = \{\texttt{closed}, \texttt{open}\}$}. We split each run in each environment of \Dreal{} in $75/25\%$ for training/testing R2SNet.

We use a Faster R--CNN~\cite{fasterrcnn} as TaskNet due to its widespread use; including its ResNet--50~\cite{resnet} backbone, it has 41\text{M} parameters. We train the TaskNet on the full \DDtwo{} and \DG{} for 60 epochs with a batch size of 4. We then deploy it for inference disabling the NMS and thresholding.

R2SNet needs to be trained from scratch on a large sample of dense proposals obtained by a TaskNet on unseen data; to do so, we augment \DDtwo{} and \DG{} as in the following.
For each image, alongside the ground--truth bounding boxes of doors, we must include the corresponding TaskNet proposals. To do this, generating these proposals using images unseen by TaskNet during its training is key, ruling out the use of the reference TaskNet trained on the full \DDtwo{} and \DG. To overcome this, we train 11 versions of Faster R--CNN, dividing the datasets into 11 segments (the first includes \DDtwo{}, the others contain images from one of the $10$ environments of \DG{}), using a leave--one--out approach. Each TaskNet, trained on $10$ segments, is used to extract proposals from the remaining, unseen, one. R2SNet is thus pre--trained with the dataset obtained by combining the 11 segments. We run a first pass training only BFNet using $\mathcal{L}_{seg}$, then in a second pass we train the whole architecture using \mbox{$\mathcal{L}_{R2S} = \mathcal{L}_{cls} + \mathcal{L}_{con\hspace{-0.1em}f} + \mathcal{L}_{sup}$} (both passes with 60 epochs and batch size of 16, $k=30$, and $W=H=32$).


The pre--trained R2SNet is adapted with data specific to its deployment environment using examples from \Dreal{}. We label this customized version as $\text{R2S}^{\#proposals}_{\#data}$, where the superscript denotes the number of proposals $k$ per training example, and the subscript indicates the percentage of data utilized relative to the total available data in \Dreal{} from the deployment environment. In particular, we assessed the impact of using $10$, $50$, and $100$ proposals, alongside $25\%$, $50\%$, and $75\%$ of the data (which correspond to $\approx 80$, $160$, and $240$ training images, respectively). The remaining $25\%$ of the data are used for testing the R2SNet, using the TaskNet as a baseline.
We apply a NMS step to the TaskNet and TaskNet+R2SNet proposals, choosing a set of conservative thresholds: \mbox{$\rho_{IoU} = 50\%$} and \mbox{$\rho_{c} = 75\%$} for the former, and \mbox{$\rho_{IoU}  = \rho_c = 50\%$} for the latter. Note that the domain adaptation of R2SNet is performed once, using the data acquired during the initial deployment of the robot, and is used later on for the whole operative life of the robot.



The performance metrics are the mean Average Precision~\cite{pascal} overall object categories (mAP) with 3 additional indicators (formally defined in~\cite{antonazzi2024development}) measuring the percentage of doors detected with the correct (wrong) label denoted as $TP$ ($FP$) and the rate between the background detections (i.e., false positives detections placed on the background) and the total number of ground truth objects, named \emph{BFD}. 
We release the implementation of R2SNet and the code to run the experiments in a publicly available repository\footnote{\url{https://aislab.di.unimi.it/research/r2snet}}.
\subsection{Results}

\begin{figure}[h]
    \centering

    \begin{subfigure}[t]{0.8\linewidth}
        \centering
\begin{tikzpicture}[scale=0.62]

\definecolor{crimson2143940}{RGB}{214,39,40}
\definecolor{darkgray176}{RGB}{176,176,176}
\definecolor{darkorange25512714}{RGB}{255,127,14}
\definecolor{forestgreen4416044}{RGB}{44,160,44}
\definecolor{lightgray204}{RGB}{204,204,204}
\definecolor{sienna1408675}{RGB}{140,86,75}
\definecolor{steelblue31119180}{RGB}{31,119,180}

\begin{axis}[
width=12cm,
height=8cm,
legend columns=4,
legend cell align={left},
legend style={
/tikz/every even column/.append style={column sep=0.3cm},
  fill opacity=0.8,
  draw opacity=1,
  text opacity=1,
  at={(0.5,0.97)},
  anchor=north,
  draw=lightgray204
},
tick align=outside,
tick pos=left,
x grid style={darkgray176},
xmin=-0.318, xmax=5.798,
xtick style={color=black},
xtick={0.44,1.59,2.74,3.89,5.04},
xticklabels={
  \(\displaystyle e_1\),
  \(\displaystyle e_2\),
  \(\displaystyle e_3\),
  \(\displaystyle e_4\),
  \(\displaystyle \overline{e}\)
},
y grid style={darkgray176},
ymin=0, ymax=79,
ytick style={color=black},
ylabel style={rotate=-90}
]
\draw[draw=black,fill=steelblue31119180,opacity=0.9,thick] (axis cs:-0.04,0) rectangle (axis cs:0.12,34.2012288763265);
\draw[draw=black,fill=steelblue31119180,opacity=0.9,thick] (axis cs:1.11,0) rectangle (axis cs:1.27,27.0618067155409);
\draw[draw=black,fill=steelblue31119180,opacity=0.9,thick] (axis cs:2.26,0) rectangle (axis cs:2.42,13.5265450015357);
\draw[draw=black,fill=steelblue31119180,opacity=0.9,thick] (axis cs:3.41,0) rectangle (axis cs:3.57,47.7587878037579);
\draw[draw=black,fill=steelblue31119180,opacity=0.9,thick] (axis cs:4.56,0) rectangle (axis cs:4.72,30.6370920992902);
\draw[draw=black,fill=forestgreen4416044,opacity=0.9,thick] (axis cs:0.16,0) rectangle (axis cs:0.32,33.5476559008558);
\draw[draw=black,fill=forestgreen4416044,opacity=0.9,thick] (axis cs:1.31,0) rectangle (axis cs:1.47,26.8614340796223);
\draw[draw=black,fill=forestgreen4416044,opacity=0.9,thick] (axis cs:2.46,0) rectangle (axis cs:2.62,19.6067470943994);
\draw[draw=black,fill=forestgreen4416044,opacity=0.9,thick] (axis cs:3.61,0) rectangle (axis cs:3.77,46.0132913217492);
\draw[draw=black,fill=forestgreen4416044,opacity=0.9,thick] (axis cs:4.76,0) rectangle (axis cs:4.92,31.5072820991567);
\draw[draw=black,fill=darkorange25512714,opacity=0.9,thick] (axis cs:0.36,0) rectangle (axis cs:0.52,49.0153175187694);
\draw[draw=black,fill=darkorange25512714,opacity=0.9,thick] (axis cs:1.51,0) rectangle (axis cs:1.67,34.7392680230501);
\draw[draw=black,fill=darkorange25512714,opacity=0.9,thick] (axis cs:2.66,0) rectangle (axis cs:2.82,25.5417119693731);
\draw[draw=black,fill=darkorange25512714,opacity=0.9,thick] (axis cs:3.81,0) rectangle (axis cs:3.97,62.2936584932592);
\draw[draw=black,fill=darkorange25512714,opacity=0.9,thick] (axis cs:4.96,0) rectangle (axis cs:5.12,42.8974890011129);
\draw[draw=black,fill=crimson2143940,opacity=0.9,thick] (axis cs:0.56,0) rectangle (axis cs:0.72,45.0875941645077);
\draw[draw=black,fill=crimson2143940,opacity=0.9,thick] (axis cs:1.71,0) rectangle (axis cs:1.87,45.0773500606389);
\draw[draw=black,fill=crimson2143940,opacity=0.9,thick] (axis cs:2.86,0) rectangle (axis cs:3.02,30.4398160392119);
\draw[draw=black,fill=crimson2143940,opacity=0.9,thick] (axis cs:4.01,0) rectangle (axis cs:4.17,64.3594410489907);
\draw[draw=black,fill=crimson2143940,opacity=0.9,thick] (axis cs:5.16,0) rectangle (axis cs:5.32,46.2410503283373);
\draw[draw=black,fill=sienna1408675,opacity=0.9,thick] (axis cs:0.76,0) rectangle (axis cs:0.92,52.552203693812);
\draw[draw=black,fill=sienna1408675,opacity=0.9,thick] (axis cs:1.91,0) rectangle (axis cs:2.07,45.6089298381157);
\draw[draw=black,fill=sienna1408675,opacity=0.9,thick] (axis cs:3.06,0) rectangle (axis cs:3.22,35.1369610750338);
\draw[draw=black,fill=sienna1408675,opacity=0.9,thick] (axis cs:4.21,0) rectangle (axis cs:4.37,74.3084102754236);
\draw[draw=black,fill=sienna1408675,opacity=0.9,thick] (axis cs:5.36,0) rectangle (axis cs:5.52,51.9016262205963);
\addplot [black, forget plot]
table {%
-0.318 0
5.798 0
};
\draw (axis cs:4.64,31.6370920992902) node[
  scale=0.8,
  anchor=south,
  text=black,
  rotate=0.0
]{31};
\draw (axis cs:4.84,32.5072820991567) node[
  scale=0.8,
  anchor=south,
  text=black,
  rotate=0.0
]{32};
\draw (axis cs:5.04,43.8974890011129) node[
  scale=0.8,
  anchor=south,
  text=black,
  rotate=0.0
]{43};
\draw (axis cs:5.24,47.2410503283373) node[
  scale=0.8,
  anchor=south,
  text=black,
  rotate=0.0
]{46};
\draw (axis cs:5.44,52.9016262205963) node[
  scale=0.8,
  anchor=south,
  text=black,
  rotate=0.0
]{52};
\end{axis}

\end{tikzpicture}
    \end{subfigure}\hfill
    \begin{subfigure}[t]{0.2\linewidth}
        \centering
        \vspace{0cm}

            \scalebox{0.7}{mAP}
            \vspace{-0.3cm}
            
            \begin{tikzpicture}[scale=0.72]

\definecolor{crimson2143940}{RGB}{214,39,40}
\definecolor{darkgray176}{RGB}{176,176,176}
\definecolor{darkorange25512714}{RGB}{255,127,14}
\definecolor{forestgreen4416044}{RGB}{44,160,44}
\definecolor{lightgray204}{RGB}{204,204,204}
\definecolor{steelblue31119180}{RGB}{31,119,180}
\definecolor{white}{RGB}{255,255,255}
\definecolor{sienna1408675}{RGB}{140,86,75}

\begin{axis}[
hide axis,
legend columns=1,
legend cell align={left},
legend style={
/tikz/every even column/.append style={column sep=0.3cm},
  fill opacity=0.8,
  draw opacity=1,
  text opacity=1,
  at={(0.5,0.95)},
  anchor=north,
  draw=white
},
xmin=-0.315, xmax=4.855,
y grid style={darkgray176},
ymin=0, ymax=130
]
\addlegendimage{ybar,area legend,draw=steelblue31119180,fill=steelblue31119180}
\addlegendentry{TaskNet}

\addlegendimage{ybar,area legend,draw=forestgreen4416044,fill=forestgreen4416044}
\addlegendentry{$\text{R2S}^{10}_{75}$}
\addlegendimage{ybar,area legend,draw=darkorange25512714,fill=darkorange25512714,opacity=0.9,thick}
\addlegendentry{$\text{R2S}^{30}_{75}$}

\addlegendimage{ybar,area legend,draw=crimson2143940,fill=crimson2143940,opacity=0.9,thick}
\addlegendentry{$\text{R2S}^{50}_{75}$}

\addlegendimage{ybar,area legend,draw=sienna1408675,fill=sienna1408675,opacity=0.9,thick}
\addlegendentry{$\text{R2S}^{100}_{75}$}
\end{axis}
\end{tikzpicture}
    \end{subfigure}
    \\\vspace{-0.6cm}
    \begin{subfigure}[t]{0.8\linewidth}
        \centering
        
        \include{images/tikz/filternet_different_boxes}
    \end{subfigure}\hfill
    \begin{subfigure}[t]{0.2\linewidth}
        \centering
        \vspace{0cm}
        \scalebox{0.7}{%
         \begin{tabular}{c}
          Additional\\ indicators
         \end{tabular}%
         } 
            \vspace{-0cm}
        \begin{tikzpicture}[scale=0.7]

\definecolor{crimson2143940}{RGB}{214,39,40}
\definecolor{darkgray176}{RGB}{176,176,176}
\definecolor{darkorange25512714}{RGB}{255,127,14}
\definecolor{forestgreen4416044}{RGB}{44,160,44}
\definecolor{lightgray204}{RGB}{204,204,204}
\definecolor{steelblue31119180}{RGB}{31,119,180}
\definecolor{white}{RGB}{255,255,255}

\begin{axis}[
hide axis,
legend columns=1,
legend cell align={left},
legend style={
/tikz/every even column/.append style={column sep=0.3cm},
  fill opacity=0.8,
  draw opacity=1,
  text opacity=1,
  at={(0.5,0.95)},
  anchor=north,
  draw=white
},
xmin=-0.315, xmax=4.855,
y grid style={darkgray176},
ymin=0, ymax=130
]

\addlegendimage{ybar,area legend,draw=black,fill=none,opacity=0.9,thick}
\addlegendentry{$TP$}
\addlegendimage{ybar,area legend,draw=black,fill=none,opacity=0.9,thick,postaction={pattern=north east lines, fill opacity=0.9}}
\addlegendentry{$FP$}
\addlegendimage{black,mark=*}
\addlegendentry{$BFD$}
\end{axis}
\end{tikzpicture}
    \end{subfigure}
    
\vspace{-1cm}
\caption{$\text{R2S}_{75}^k$ performance when varying $k$ expressed with the mAP (top row) and the additional indicators (bottom row).}
\label{fig:proposals_increasing}
\vspace{-0.3cm}
\end{figure}
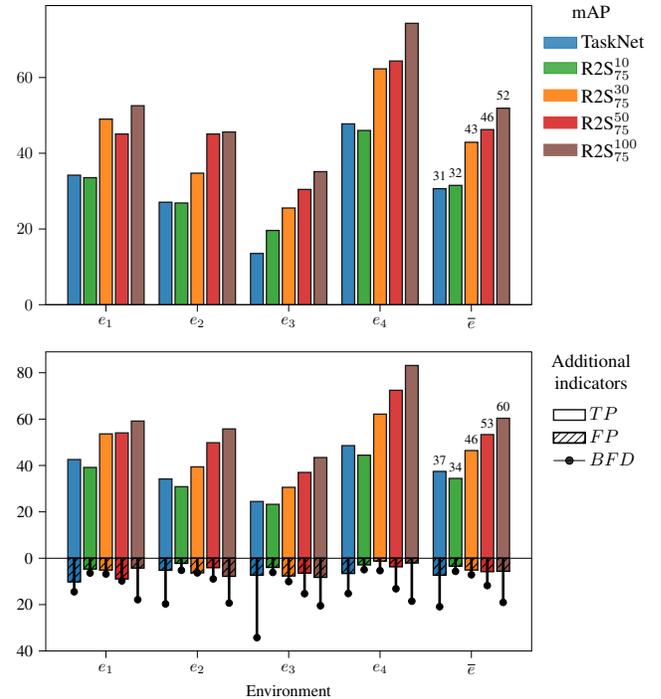

Fig.~\ref{fig:proposals_increasing} shows the performance of $\text{R2S}_{75}^k$ varying the number of proposals $k$. A value $k \in \{30,50\}$ improves the mAP of $\approx 45\%$, while also reducing both $FP$ and $BFD$, showing the effectiveness of our R2SNet. Despite the filtering becomes challenging due to the high number of noisy low--confidence proposals, a higher value of $k=100$ results in a further increase of the mAP, at the expense of higher $BFD$ (that are still close to those obtained by the TaskNet).  Interestingly, also a low value of $k=10$ has some positive aspects; while it does not improve the mAP, it still reduces the $FP$ and $BFD$. Fig.~\ref{fig:proposals_increasing} shows the benefits brought by the $\text{R2S}_{75}^{100}$ refinement to the TaskNet raw proposals.

Another interesting remark, confirming the solidity of our approach, comes from evaluating the results when a different and increasing number of samples obtained in the robot target environment are used to train R2SNet, as shown in Table~\ref{tab:data_quantity}. For this test, we used $k=30$. Even with a few examples, $\text{R2S}^{30}_{25}$ increases both the mAP and $TP$ of $\approx 20\%$ while halving the percentage of $BFD$. Of course, exposing the R2SNet to a higher number of examples, as in $\text{R2S}^{30}_{50-75}$, increases performances, but our findings suggest that a few examples are enough for R2SNet to improve the TaskNet performance on the target environment.


To further evaluate the contribution of each network's component, we conduct an ablation study where the relabeling, rescoring, and refinement heads are incrementally activated. To do this, we use as a reference $\text{R2S}_{75}^{100}$, which has the best performances. Table~\ref{tab:ablation} reports the results averaged over the 4 environments.
From this analysis, it can be seen how the relabeling head alone ensures a significant mAP and $TP$ improvement while reducing the $FP$ and $BFD$. The use of the rescoring head similarly improves the mAP and $TP$, without reducing $FP$ and $BFD$s. The best overall performance is shown when all three heads are used. However, the suppression head has less impact on the performance than the other two heads; still, it is needed for training, as disabling it increases the $BFD$ of the $3\%$.

\begin{table}[h]

\centering
\begin{tabular}{ccc|c@{\hskip 0.2cm}c@{\hskip 0.2cm}c@{\hskip 0.2cm}c@{\hskip 0.1cm}}
\toprule
         & & & \multicolumn{4}{c}{$\overline{e}$} \\
         \textbf{Rel.} &\textbf{Res.} &\textbf{Sup.} & \textbf{mAP}$\uparrow$& $\mathbf{TP}$$\uparrow$ &  $\mathbf{FP}$$\downarrow$ & $\mathbf{BFD}$$\downarrow$  \\
\midrule
 \ & \ & \ &34 &44\% & 10\%  & 35\% \\
 \Vmark & \ & \ &44 &48\% & 4\%  & 6\% \\
 \ & \Vmark & \ &41 &54\% & 15\%  & 34\% \\
  \ & \ & \Vmark &37 &43\% & 9\%  & 14\% \\
 \Vmark & \Vmark & \ &52 &61\% & 6\%  & 20\% \\

 \Vmark & \ & \Vmark &44 &47\% & 4\%  & 5\% \\
 \ & \Vmark & \Vmark &41 &53\% & 15\%  & 31\% \\
 \Vmark & \Vmark & \Vmark &52 &60\% & 6\%  & 19\% \\
\bottomrule
\end{tabular}
\caption{Ablation study results.}\label{tab:ablation}
\vspace{-0.5cm}
\end{table}

Our R2SNet has 8M parameters. To assess its computational demands for inference, we installed it on an edge device, an NVIDIA Jetson TX2, mounted on the Giraff--X robot. On this hardware, commonly available in service robots, R2SNet demonstrates remarkable efficiency, processing images at a rate of 16.7 Hz on the GPU and 2.6 Hz on the CPU. As a reference, TaskNet processes at significantly lower frequencies of 1.1 Hz and 0.06 Hz, respectively, on the same platforms. The R2SNet training with a RTX 3090 GPU took a few minutes.
These results show how our method can be used in real--time on a mobile robot, illustrating R2SNet's capability for efficient deployment in robots that utilize edge devices, even those without a GPU.
\section{Conclusions}

This paper presented a scalable domain adaptation solution for robotic ecosystems relying on cloud--based object detection. We propose R2SNet, a novel lightweight architecture to refine the proposals locally in the robot to mitigate the performance degradation caused by domain shifts.

In future work, we will validate our results with other TaskNet architectures and extend our approach enabling the robots to upload domain--independent images to enhance the TaskNet performance. Furthermore, we plan to investigate techniques for adaptive learning for evolving environments.

\section*{Acknowledgements}
This work was partially funded by Grant Number G53D23002860006 - PRIN2022, by the Italian Ministery of Research and University.

\bibliographystyle{IEEEtran}
\bibliography{./contents/citations}

\begin{thebibliography}{10}
\providecommand{\url}[1]{#1}
\csname url@samestyle\endcsname
\providecommand{\newblock}{\relax}
\providecommand{\bibinfo}[2]{#2}
\providecommand{\BIBentrySTDinterwordspacing}{\spaceskip=0pt\relax}
\providecommand{\BIBentryALTinterwordstretchfactor}{4}
\providecommand{\BIBentryALTinterwordspacing}{\spaceskip=\fontdimen2\font plus
\BIBentryALTinterwordstretchfactor\fontdimen3\font minus \fontdimen4\font\relax}
\providecommand{\BIBforeignlanguage}[2]{{%
\expandafter\ifx\csname l@#1\endcsname\relax
\typeout{** WARNING: IEEEtran.bst: No hyphenation pattern has been}%
\typeout{** loaded for the language `#1'. Using the pattern for}%
\typeout{** the default language instead.}%
\else
\language=\csname l@#1\endcsname
\fi
#2}}
\providecommand{\BIBdecl}{\relax}
\BIBdecl

\bibitem{alatise2020review}
M.~B. Alatise and G.~P. Hancke, ``A review on challenges of autonomous mobile robot and sensor fusion methods,'' \emph{IEEE Access}, vol.~8, pp. 39\,830--39\,846, 2020.

\bibitem{oza2023unsupervised}
P.~Oza, V.~A. Sindagi, V.~V. Sharmini, and V.~M. Patel, ``Unsupervised domain adaptation of object detectors: A survey,'' \emph{IEEE Trans. Pattern Anal. Mach. Int.}, 2023.

\bibitem{lee2022surgical}
Y.~Lee, A.~S. Chen, F.~Tajwar, A.~Kumar, H.~Yao, P.~Liang, and C.~Finn, ``Surgical fine-tuning improves adaptation to distribution shifts,'' in \emph{Proc. ICLR}, 2023.

\bibitem{hu2012cloud}
G.~Hu, W.~P. Tay, and Y.~Wen, ``Cloud robotics: architecture, challenges and applications,'' \emph{IEEE Netw.}, vol.~26, no.~3, pp. 21--28, 2012.

\bibitem{tanwani2019fog}
A.~K. Tanwani, N.~Mor, J.~Kubiatowicz, J.~E. Gonzalez, and K.~Goldberg, ``A fog robotics approach to deep robot learning: Application to object recognition and grasp planning in surface decluttering,'' in \emph{Proc. {ICRA}}.\hskip 1em plus 0.5em minus 0.4em\relax IEEE, 2019, pp. 4559--4566.

\bibitem{hosang-learning-nms}
J.~Hosang, R.~Benenson, and B.~Schiele, ``Learning non-maximum suppression,'' in \emph{Proc. CVPR}, July 2017.

\bibitem{antonazzi2024development}
M.~Antonazzi, M.~Luperto, N.~A. Borghese, and N.~Basilico, ``Development and adaptation of robotic vision in the real-world: the challenge of door detection,'' 2024.

\bibitem{antonazzi2023enhancing}
M.~Antonazzi, M.~Luperto, N.~Basilico, and N.~A. Borghese, ``Enhancing door-status detection for autonomous mobile robots during environment-specific operational use,'' in \emph{Proc. ECMR}, 2023.

\bibitem{afrin2021resource}
M.~Afrin, J.~Jin, A.~Rahman, A.~Rahman, J.~Wan, and E.~Hossain, ``Resource allocation and service provisioning in multi-agent cloud robotics: A comprehensive survey,'' \emph{IEEE Commun. Surv. Tutor.}, vol.~23, no.~2, pp. 842--870, 2021.

\bibitem{liu2023roboec2}
B.~Liu, L.~Wang, and M.~Liu, ``{RoboEC2}: A novel cloud robotic system with dynamic network offloading assisted by amazon {EC2},'' \emph{IEEE T AUTOM SCI ENG}, pp. 1--15, 2023.

\bibitem{wang2020convergence}
X.~Wang, Y.~Han, V.~C. Leung, D.~Niyato, X.~Yan, and X.~Chen, ``Convergence of edge computing and deep learning: A comprehensive survey,'' \emph{IEEE Commun. Surv. Tutor.}, vol.~22, no.~2, pp. 869--904, 2020.

\bibitem{guo2019distributed}
Y.~Guo, B.~Zou, J.~Ren, Q.~Liu, D.~Zhang, and Y.~Zhang, ``Distributed and efficient object detection via interactions among devices, edge, and cloud,'' \emph{IEEE Trans. Multimed.}, vol.~21, no.~11, pp. 2903--2915, 2019.

\bibitem{abuadbba2020can}
S.~Abuadbba, K.~Kim, M.~Kim, C.~Thapa, S.~A. Camtepe, Y.~Gao, H.~Kim, and S.~Nepal, ``Can we use split learning on 1d cnn models for privacy preserving training?'' in \emph{Proc. ASIACCS}, 2020, pp. 305--318.

\bibitem{chakroun2021distributing}
I.~Chakroun, T.~Vander~Aa, R.~Wuyts, and W.~Verachtert, ``Distributing intelligence for object detection using edge computing,'' in \emph{Proc. CLOUD}.\hskip 1em plus 0.5em minus 0.4em\relax IEEE, 2021, pp. 681--687.

\bibitem{teerapittayanon2017distributed}
S.~Teerapittayanon, B.~McDanel, and H.-T. Kung, ``Distributed deep neural networks over the cloud, the edge and end devices,'' in \emph{Proc. ICDCS}.\hskip 1em plus 0.5em minus 0.4em\relax IEEE, 2017, pp. 328--339.

\bibitem{FogROS2}
J.~Ichnowski, K.~Chen, K.~Dharmarajan, S.~Adebola, M.~Danielczuk \emph{et~al.}, ``Fogros2: An adaptive platform for cloud and fog robotics using ros 2,'' in \emph{Proc. ICRA}, 2023, pp. 5493--5500.

\bibitem{tanwani2020rilaas}
A.~K. Tanwani, R.~Anand, J.~E. Gonzalez, and K.~Goldberg, ``Rilaas: Robot inference and learning as a service,'' \emph{IEEE RA-L}, vol.~5, no.~3, pp. 4423--4430, 2020.

\bibitem{vinod2022development}
D.~Vinod and P.~SaiKrishna, ``Development of an autonomous fog computing platform using control-theoretic approach for robot-vision applications,'' \emph{ROBOT AUTON SYST}, vol. 155, p. 104158, 2022.

\bibitem{liu2016ssd}
W.~Liu, D.~Anguelov, D.~Erhan, C.~Szegedy, S.~Reed, C.-Y. Fu, and A.~C. Berg, ``Ssd: Single shot multibox detector,'' in \emph{Proc. ECCV}.\hskip 1em plus 0.5em minus 0.4em\relax Springer, 2016, pp. 21--37.

\bibitem{beksi2015core}
W.~J. Beksi, J.~Spruth, and N.~Papanikolopoulos, ``Core: A cloud-based object recognition engine for robotics,'' in \emph{Proc. IROS}.\hskip 1em plus 0.5em minus 0.4em\relax IEEE, 2015, pp. 4512--4517.

\bibitem{penmetcha2020smart}
M.~Penmetcha, S.~S. Kannan, and B.-C. Min, ``Smart cloud: Scalable cloud robotic architecture for web-powered multi-robot applications,'' in \emph{Proc. SMC}.\hskip 1em plus 0.5em minus 0.4em\relax IEEE, 2020, pp. 2397--2402.

\bibitem{chinchali2021network}
S.~Chinchali, A.~Sharma, J.~Harrison, A.~Elhafsi, D.~Kang, E.~Pergament, E.~Cidon, S.~Katti, and M.~Pavone, ``Network offloading policies for cloud robotics: a learning-based approach,'' \emph{Auton. Robot.}, vol.~45, no.~7, pp. 997--1012, 2021.

\bibitem{objectdetectionsurvey}
Z.~Zou, K.~Chen, Z.~Shi, Y.~Guo, and J.~Ye, ``Object detection in 20 years: A survey,'' \emph{Proc. IEEE}, vol. 111, no.~3, pp. 257--276, 2023.

\bibitem{fasterrcnn}
S.~Ren, K.~He, R.~Girshick, and J.~Sun, ``Faster {R-CNN}: Towards real-time object detection with region proposal networks,'' \emph{Adv. Neur. In.}, vol.~28, no.~6, 2015.

\bibitem{yolov3}
A.~Farhadi and J.~Redmon, ``Yolo{V}3: An incremental improvement,'' 2018.

\bibitem{popordanoska_calibration_od}
T.~Popordanoska, A.~Tiulpin, and M.~B. Blaschko, ``Beyond classification: Definition and density-based estimation of calibration in object detection,'' in \emph{Proc. WACV}, January 2024, pp. 585--594.

\bibitem{pointnet}
C.~R. Qi, H.~Su, K.~Mo, and L.~J. Guibas, ``Pointnet: Deep learning on point sets for 3d classification and segmentation,'' in \emph{Proc. CVPR}, July 2017.

\bibitem{resnet}
K.~He, X.~Zhang, S.~Ren, and J.~Sun, ``Deep residual learning for image recognition,'' in \emph{Proc. CVPR}, 2016, pp. 770--778.

\bibitem{fpn}
T.-Y. Lin, P.~Dollar, R.~Girshick, K.~He, B.~Hariharan, and S.~Belongie, ``Feature pyramid networks for object detection,'' in \emph{Proc. CVPR}, July 2017.

\bibitem{deepdoors2}
J.~Ramôa, V.~Lopes, L.~Alexandre, and S.~Mogo, ``Real-time 2d–3d door detection and state classification on a low-power device,'' \emph{{SN Appl. Sci.}}, 2021.

\bibitem{gibson}
F.~Xia, A.~R. Zamir \emph{et~al.}, ``Gibson env: Real-world perception for embodied agents,'' in \emph{Proc. CVPR)}, 2018.

\bibitem{giraff}
M.~Luperto, M.~Romeo, J.~Monroy, J.~Renoux, A.~Vuono \emph{et~al.}, ``User feedback and remote supervision for assisted living with mobile robots: A field study in long-term autonomy,'' \emph{ROBOT AUTON SYST}, vol. 155, p. 104170, 2022.

\bibitem{pascal}
M.~Everingham, L.~V. Gool, C.~K.~I. Williams, J.~M. Winn, and A.~Zisserman, ``The pascal visual object classes ({VOC}) challenge,'' \emph{Int. J. Comput. Vision}, vol.~88, pp. 303--338, 2009.

\end{thebibliography}

\end{document}